\documentclass[letterpaper, 10 pt, conference]{ieeeconf}  

\IEEEoverridecommandlockouts                              

\usepackage[ruled,vlined,linesnumbered]{algorithm2e}
\usepackage{amsmath, graphicx}
\usepackage{balance}
\usepackage{amssymb,amsbsy}
\usepackage{breqn,bbm,xcolor}
\usepackage{multirow}
\usepackage[symbol,bottom]{footmisc}
\usepackage{soul}
\usepackage{url}
\usepackage{booktabs} 
\usepackage{tikz}
\usetikzlibrary{positioning}
\usetikzlibrary{arrows}
\usetikzlibrary{shapes,snakes}
\usetikzlibrary{plotmarks}
\usepackage{pgfplots}
\pgfplotsset{compat=1.15}
\usepackage[caption=false,font=footnotesize]{subfig}
\usepackage{dblfloatfix}
\usepackage{esvect}

\DeclareUnicodeCharacter{2212}{-}

\DeclareMathOperator*{\argmin}{arg\,min}

\title{\LARGE \bf
Image-based Pose Estimation and Shape Reconstruction for Robot Manipulators and Soft, Continuum Robots via Differentiable Rendering
}

\author{Jingpei Lu$^{1,\dagger}$, Fei Liu$^{1,\dagger}$, C\'edric Girerd$^{1,2}$, and Michael C. Yip$^{1}$ 
\thanks{This project was funded by the US Army Telemedicine and Advanced Technologies Research Center and NSF Awards \#2045803 and \#1935329.}
\thanks{$\dagger$ These authors contributed equally.}
\thanks{$^{1}$Department of Electrical and Computer Engineering, University of California San Diego, La Jolla, CA 92093 USA.{\tt\small\{jil360, f4liu, cgirerd, yip\}@ucsd.edu}}
\thanks{$^{2}$LIRMM, Univ Montpellier, CNRS, Montpellier, France. {\tt\small{cedric.girerd@lirmm.fr}}. This work was conducted while he was a visiting scholar at UCSD.}}

\begin{document}

\maketitle

\begin{abstract}
State estimation from measured data is crucial for robotic applications as autonomous systems rely on sensors to capture the motion and localize in the 3D world. 
Among sensors that are designed for measuring a robot's pose, or for soft robots, their shape, vision sensors are favorable because they are information-rich, easy to set up, and cost-effective.
With recent advancements in computer vision, deep learning-based methods no longer require markers for identifying feature points on the robot. 
However, learning-based methods are data-hungry and hence not suitable for soft and prototyping robots, as building such bench-marking datasets is usually infeasible.
In this work, we achieve image-based robot pose estimation and shape reconstruction from camera images. Our method requires no precise robot meshes, but rather  utilizes a differentiable renderer and primitive shapes. It hence can be applied to robots for which CAD models might not be available or are crude. Our parameter estimation pipeline is fully differentiable. The robot shape and pose are estimated iteratively by back-propagating the image loss to update the parameters. We demonstrate that our method of using geometrical shape primitives can achieve high accuracy in shape reconstruction for a soft continuum robot and pose estimation for a robot manipulator. 
\end{abstract}

\section{Introduction}

Sensory feedback of state parameters, such as the position and body configuration of a robot in its environment, is a fundamental requirement for operating autonomous systems in real-world, unknown spaces. In-place sensing may exist with motor encoders for robot manipulators, or Fiber Braggs for soft robots~\cite{shi2016shape}, all providing an estimate of their relative pose and body configurations in relation to the real world environment; however, the limitation is they all exhibit cumulative position errors due to long kinematic chains. In addition, for both soft and rigid robots, the procedure for mounting internal sensors can be tricky and wiring and communication lines can constrain the mechanics and articulation of the robot. That is why measuring the body configuration of a soft robot is notoriously challenging. 

If the goal is to observe and track the motion of robots in the wild, e.g., for behavior cloning or offline reinforcement learning, the robot's state information may not be readily available or even accessible. A good example is in minimally invasive surgery (MIS), where over 1 million procedures are performed yearly on a daVinci Surgical Robot, many researchers are interested in automating aspects of the surgical procedures~\cite{yip2019robot}. Video of surgeons all over the world using the same robot to perform tasks like suturing, where properly grasping suture needles is an expert skill for which data could be useful for learning control policies as shown in~\cite{chiu2021bimanual}. However, these datasets only have video data and lack kinematic data from the robot due to proprietary access. 
\begin{figure}[t]
\vspace{0mm}
    \centering
    \includegraphics[width=\linewidth]{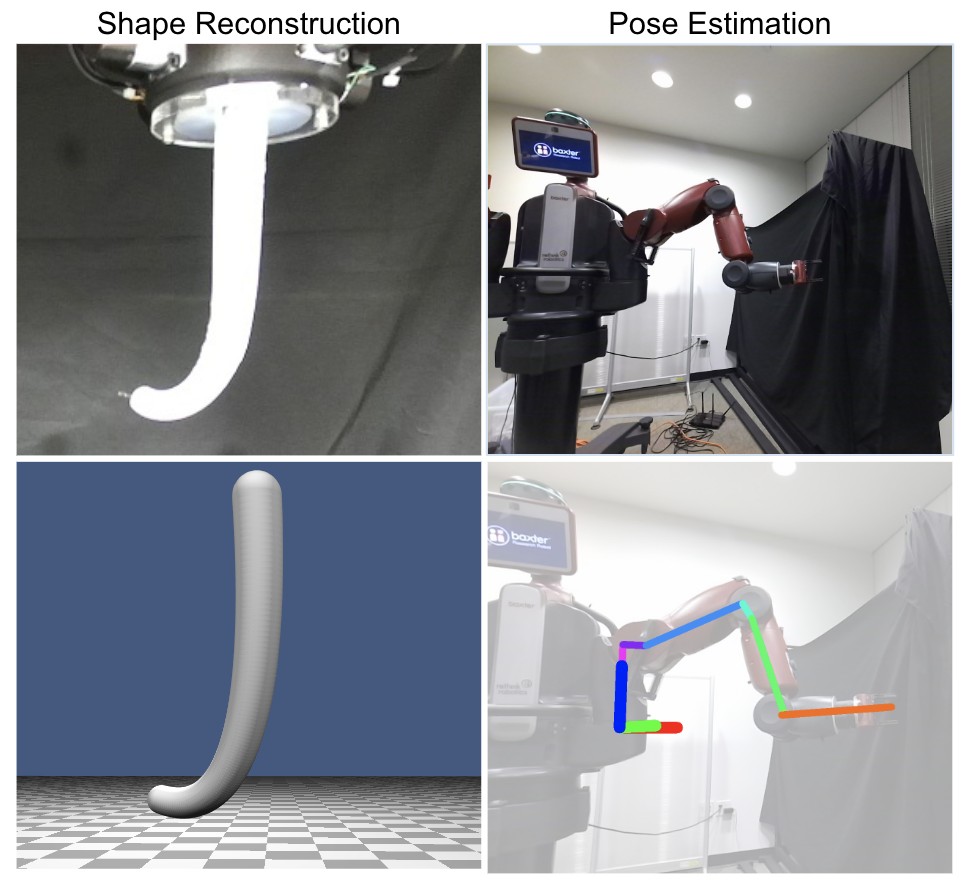}
    \vspace{-7mm}
    \caption{Robot shape reconstruction and pose estimation via differentiable rendering. The top row shows the real images. The bottom row shows the estimated robot shape (left) and pose (right).}
    \label{fig:cover}
\vspace{-0.2in}
\end{figure}

\begin{figure*}[b!]
    \vspace{-0.1in}
    \centering
    \includegraphics[width=.9\linewidth]{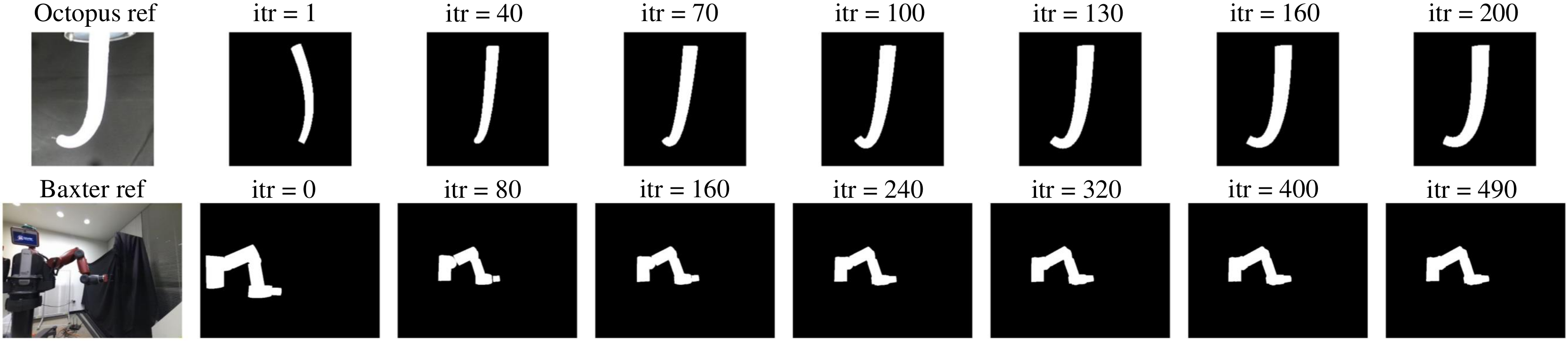}
    \caption{ \textcolor{black}{Visualization of the optimization process. We show the rendered silhouettes at different iterations to demonstrate the convergence of our algorithm.}
    }
    \label{fig:optimization_sequence_img}
\end{figure*}

Ultimately, in the above MIS and many other \textit{in-the-wild} robot scenarios, tracking the robot pose and body configurations \textit{directly from a camera} offers the greatest flexibility. They are easy to set up or are already recording, do not require access to internal robot sensors, are affordable and widely ubiquitous. Traditionally, fiducial markers, like ArUco marker~\cite{garrido2014aruco} and AprilTag~\cite{olson2011apriltag}, are widely used for robot pose estimation. These markers are attached to the specific locations of the robot and the robot state parameters can be estimated by knowing the kinematics model. However in most unstructured environments, it is unrealistic to have these markers attached; furthermore, in dirty environments like MIS or constrained environments, these markers can be permanently obscured or occluded. For soft robot applications, their body deformations and their tendency for full-body contact with environments and objects make it frequently impractical for securing fiducials or template markers.

The most flexible way to estimate pose from a camera is to do marker-less tracking. With recent advancements in Computer Vision, Convolutional Neural Networks (CNN) present a promising way for marker-less feature detection~\cite{lee2020camera}\cite{lu2022pose} which no longer requires physical modification on the robot. In spite of the success of the CNN, training a neural network requires significant amounts of labeled datasets which is usually infeasible for soft robots and other robot prototypes, or where labeling of data is costly (e.g., MIS).
Recently, in computer graphics, differentiable rendering has proved to be effective in image-based reconstruction by computing the derivative of images with respect to scene parameters such as camera pose and object geometry~\cite{kato2018neural}\cite{kato2020differentiable}\cite{liu2019soft}. This could be translated to the task of marker-less pose tracking.

In this paper, we demonstrate the capability of estimating robot pose and configuration directly from a camera, as shown in Fig.~\ref{fig:cover}. The method works via the technique of differentiable rendering, and can be effective both in rigid-link robot manipulators as well as soft continuum robots. This is uniquely challenging, as soft continuum robots have an infinite number of configurations, while some rigid robot manipulators may not come with predefined CAD models for their users (these are typically proprietary). Under these constraints, we are still able to estimate a robot's state without knowledge of a high-resolution CAD by introducing a flexible method involving shape primitives that work across a wide range of robots.
Our contributions are:
\begin{enumerate}
    \item We propose a general framework for parameter estimation by utilizing differentiable rendering with geometrical shape primitives.
    \item \textcolor{black}{The framework generalizes to both rigid and soft continuum robots for parameter estimation.}
    \item We investigate the novel loss functions to overcome the local minima when applying differentiable rendering to the objective of robot pose estimation.
\end{enumerate}
To examine the effectiveness of our framework, we collect a synthetic and a real dataset for a soft continuum robot and reconstruct the robot shape by estimating the curve parameters. We also evaluate our method on robot pose estimation where the 6~Degree-of-Freedom (DOF) camera-to-robot pose for a Baxter robot is estimated with provided RGB images and joint encoder readings. The experimental results show that our method is out-performing the state-of-the-art pose estimation algorithms.

\section{Previous Work}
A few techniques in image-based estimation of robot poses and shapes have been previously explored.

\textit{Image-based Shape Reconstruction for Soft Continuum Robots}:
Image-based measurements of soft continuum robots are very task-specific and only work in certain environments. 
Techniques include using fluoroscopy~\cite{papalazarou20123d,hoffmann2012semi} and ultrasound~\cite{waine20153d, cheung2004enhancement}.
These image-based techniques require specific imaging sources which might not always be available.
\cite{AlBeladi_2021, Cabras_2017, Reilink_2012} consider using endoscopic images for shape reconstruction while the markers are required for identifying predefined feature points. Moreover,~\cite{Croom_2010, Shan_2018} also introduce the shape reconstruction methods of using stereo images and depth data. In contrast, we will be focusing on markerless shape reconstruction from a single RGB camera.

\textit{Image-based Robot Pose Estimation}:
The common approach for image-based robot pose estimation is to attach the fiducial markers~\cite{garrido2014aruco,olson2011apriltag} to known locations along the robot kinematic chain. Given the joint angles, the position of the marker in the robot base frame can be calculated and the robot pose can be derived by solving an optimization problem~\cite{li2020super,ilonen2011robust,richter2021robotic}.
More recently, deep learning brings a promising way of marker-less pose estimation, where a CNN is trained to extract predefined feature points and the robot pose is estimated by solving the Perspective-n-Point problem~\cite{lambrecht2019towards,lee2020camera,lu2021super,lu2022pose}.
Meanwhile, rendering-based methods also demonstrate their advantages in robot pose estimation by using the high-resolution robot CAD model for more precise estimation~\cite{hao2018vision,labbe2021single}.
Our work utilizes differentiable rendering which requires no robot CAD model or large dataset for model training.

\section{Methodology}
We consider the problem of estimating the robot state parameters $\Theta$ from a single RGB image. Specifically, we estimate the robot pose and configurations by minimizing differences between the observed RGB image and a rendered reconstruction image. This is formulated as follows:
\begin{equation}
    \Theta^* = \argmin_{\Theta} \mathcal{L} (f_{mask}(\mathbb{I}), f_{render}(\Theta))
\end{equation}
where $f_{mask}$ processed the given RGB image $\mathbb{I}$ into a binarized mask image for the robot. The function $f_{render}$ takes in the \textcolor{black}{estimated parameters}, reconstructs \textcolor{black}{the robot mesh}, and renders the reconstruction. We aim to estimate the state parameters by minimizing the objective loss function $\mathcal{L}$. A visual of the final optimization process are shown in Fig.~\ref{fig:optimization_sequence_img}. The process to get to this stage is described below.


\subsection{The State Parameter Estimation Framework}

 \begin{algorithm}[t]
    \caption{Robot State Parameter Estimation via Differentiable Rendering}
    \label{alg:main_outline}
    \SetKwInOut{Input}{Input}
    \SetKwInOut{Output}{Output}
    \Input{Image frame $\mathbb{I}$, initialization $\Theta_{state}^{(0)}, \Theta_{verts}^{(0)}$}
    \Output{Estimated robot state parameters $\Theta_{state}^*$}
    \tcp{Generate robot masks}
    $\mathbb{M}^{ref} \leftarrow f_{mask}(\mathbb{I})$\\

    \tcp{Optimization loop}
    $\mathcal{L}_{min} = \infty$\\
    \For{$i = 0$ \KwTo $N_o$}{
        \tcp{Section III-B}
        $\mathcal{M}^{(i)} \leftarrow reconstructMesh(\Theta_{state}^{(i)}, \Theta_{verts}^{(i)})$\\
        \tcp{Section III-C}
        $\mathbb{S}^{(i)} \leftarrow silhouetteRendering(\mathcal{M}^{(i)})$\\
        $\mathcal{L}^{(i)} \leftarrow computeLoss(\mathbb{S}^{(i)},\mathbb{M}^{ref})$\\
        
        \If{$\mathcal{L}^{(i)} < \mathcal{L}_{min}$}{
            $\mathcal{L}_{min} = \mathcal{L}^{(i)}$ \\
            $\Theta_{state}^* = \Theta_{state}^{(i)}$\\
        }
        $\Theta_{verts}^{(i+1)} = \Theta_{verts}^{(i)} - \lambda_{verts} \frac{\partial \mathcal{L}^{(i)}}{\partial \Theta_{verts}^{(i+1)}}$\\
        $\Theta_{state}^{(i+1)} = \Theta_{state}^{(i)} - \lambda_{state} \frac{\partial \mathcal{L}^{(i)}}{\partial \Theta_{state}^{(i+1)}}$\\
    }

    \Return{ $\Theta_{state}^*$ }
\end{algorithm}

The overall framework for state parameter estimation is described in the Algorithm~\ref{alg:main_outline}. 
We first process the observed RGB image $\mathbb{I}$ into a binary mask $\mathbb{M}^{ref}$, which segments the robot pixel from the background. The binary mask contains value 1 for the pixels that belong to the robot and 0 otherwise. \textcolor{black}{In our implementation,} the segmentation is achieved by color segmentation for the soft continuum robot (Section~\ref{section:shape_reconstruction}) and a CNN-based semantic segmentation for the robot manipulator (Section~\ref{section:pose_estimation}).
We also initialize a robot mesh in a renderer as a set of geometrical primitive shapes with predefined vertices, edges, and faces. 

During the iterative optimization process, we estimate the deformation of mesh vertices parameterized by $\Theta_{verts}$ and reconstruct the robot mesh with state parameters $\Theta_{state}$ (Section~\ref{methods:reconstruct_robot_mesh}). 
We render a silhouette image $\mathbb{S}$ from the reconstruction and compare it with the reference masked image $\mathbb{M}^{ref}$. A loss $\mathcal{L}$ is computed based on the curated objective functions (Section~\ref{method:diff_rendering}). Since the full reconstruction and rendering pipeline is differentiable, a gradient on the loss may be taken with respect to the parameters and the objectives can be optimized (lines 11-12 in Algorithm 1). We iterate the optimization process for $N_o$ times and output the state parameters that minimize the objective loss.

\subsection{Reconstruct Robot Mesh with Geometric Primitives}
\label{methods:reconstruct_robot_mesh} 

In this section, we describe the methods of reconstructing the robot mesh using geometric primitives for the soft and rigid robot, respectively. Note that state parameters $\Theta_{state}$ and mesh vertex parameters $\Theta_{verts}$ are defined differently according to their body types.  

\vspace{2mm}
\subsubsection{Mesh Reconstruction for Soft Continuum Robot}
\label{methods:soft_robot}


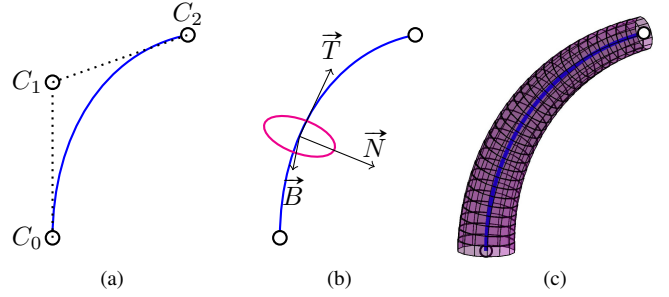
\begin{figure}[t!]
\subfloat[]{\begin{tikzpicture}[scale=0.9]
    \draw[thick, blue, -] (0, 0) .. controls (0, 1.5) and (0.75, 2.75) .. (2, 3);
    \draw[thick, black, fill=white] (0, 0) circle (1mm) node[anchor=east] {$C_0$};
    \draw[thick, black, fill=white] (0, 2.3) circle (1mm) node[anchor=east] {$C_1$};
    \draw[thick, black, fill=white] (2, 3) circle (1mm) node[anchor=south] {$C_2$};
    \draw[thick, black, dotted] (0, 0) -- (0, 2.3);
    \draw[thick, black, dotted] (0, 2.3) -- (2, 3);
\end{tikzpicture}\label{fig:bezier_a}}
\hfill
\subfloat[]{\begin{tikzpicture}[scale=0.9]
    \draw[thick, blue, -] (0, 0) .. controls (0, 1.5) and (0.75, 2.75) .. (2, 3);
    \draw[rotate around={-20:(0.29,1.5)}, magenta, thick] (0.29,1.5) ellipse (0.55cm and 0.25cm);
    \draw[thick, black, fill=white] (0, 0) circle (1mm) node[anchor=east] {\textcolor{white}{$C_0$}};
    \draw[thick, black, fill=white] (2, 3) circle (1mm) node[anchor=south] {\textcolor{white}{$C_2$}};
    \draw[->](0.29,1.5)--(0.76,2.5) node[anchor=south] {$\vv{T}$};
    \draw[->](0.29,1.5)--(1.4,1.05) node[anchor=south] {$\vv{N}$};
    \draw[->](0.29,1.5)--(0.19,1) node[anchor=north] {$\vv{B}$};
\end{tikzpicture}\label{fig:bezier_b}}
\hfill
\subfloat[]{\includegraphics[width=0.30\linewidth]{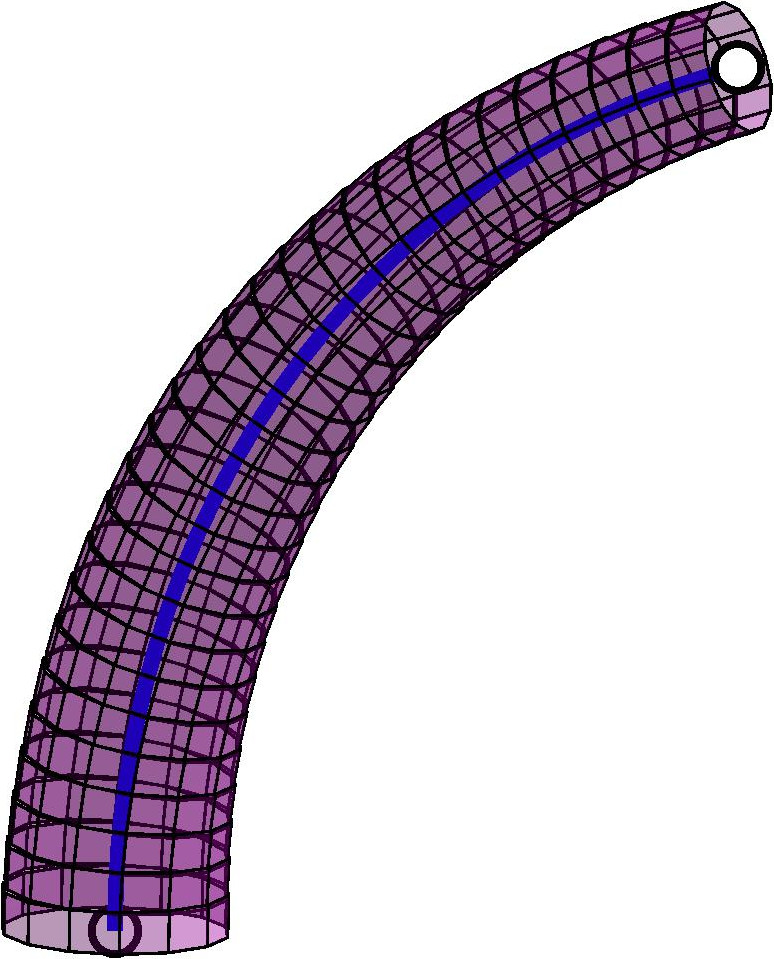}\label{fig:bezier_c}}
\caption{\label{fig:contruction_surface_mesh_soft_robot} Illustration of surface mesh construction for soft continuum robot, with \protect\subref{fig:bezier_a}~the B\'{e}zier curve and control points, \protect\subref{fig:bezier_b}~the Frenet–Serret frame of cross section, and \protect\subref{fig:bezier_c}~the constructed surface mesh.}
\vspace{-0.14in}
\end{figure}

The shape of a soft continuum robot can be described in several ways, most easily using a constant curvature model~\cite{Webster_2010}. However, since this is a limiting approximation, instead a better model chosen is a B\'{e}zier curve model, which expresses a \textcolor{black}{smooth and continuous} curve with arbitrary curvature in 3D space. A Given a set of $N$ control points $\{\mathbf{c}_{i} | \mathbf{c}_{i} \in \mathbb{R}^3 \}^{N}_{i=0}$, the shape of the curve is defined as:
\begin{equation}
    \mathbf{p}(s) = \sum_{i=0}^N \frac{N!}{i!(N-i)!}(1-s)^{N-i} s^i \mathbf{c}_i \; , \; 0 \leq s \leq 1,
\end{equation}
For simplicity, we use a quadratic B\'{e}zier curve ($N=2$) and estimate the state of the control points $\Theta_{state}$ (see Fig.~\ref{fig:bezier_a}).

\textcolor{black}{
In general, the surface mesh for a continuum robot can be approximated as the tubular structure~\cite{Shiyao_2021}. A tubular surface is defined as a union of cross sections, and each cross-section is centered at the axis along the 3D curve, as shown in Fig~\ref{fig:contruction_surface_mesh_soft_robot}.
To describe 3D coordinate frames along a quadratic B\'{e}zier curve, we compute the Frenet–Serret frame which is defined by a unit vector $\mathbf{T}$ tangent to the curve, a unit vector $\mathbf{N}$ normal to the curve, and a unit vector $\mathbf{B}$ perpendicular to the tangent and normal vectors (Fig.~\ref{fig:bezier_b}).
The Frenet–Serret coordinates, parameterized by $s$, are defined as:}
\begin{equation}
\begin{split}
    & \mathbf{T}(s) = \frac{\mathbf{p}'(s)}{\left\| \mathbf{p}'(s) \right\| } \\
    & \mathbf{N}(s) = \frac{\mathbf{T}'(s)}{\left\|  \mathbf{T}'(s) \right\| } = \frac{\mathbf{p}'(s) \times \left(\mathbf{p}''(s) \times \mathbf{p}'(s) \right)}{ \left\|\mathbf{p}'(s)\right\| \left\|\mathbf{p}''(s) \times \mathbf{p}'(s)\right\| }\\
    & \mathbf{B}(s) = \mathbf{T}(s) \times \mathbf{N}(s) = \frac{\mathbf{p}'(s)\times \mathbf{p}''(s)}{ \left\|\mathbf{p}'(s)\times \mathbf{p}''(s)\right\| }\\
\end{split}
\end{equation}
\textcolor{black}{where $\mathbf{p}'(s), \mathbf{p}''(s)$ are the first and second derivatives of the quadratic B\'{e}zier curve model:}
\begin{equation}
\begin{split}
    & \mathbf{p}(s) = (1-s)^2 \mathbf{c}_0 + 2(1-s)s \mathbf{c}_1 + s^2 \mathbf{c}_2\\
    & \mathbf{p}'(s)=2(1-s)(\mathbf{c}_{1}-\mathbf{c}_{0})+2s(\mathbf{c}_{2}-\mathbf{c}_{1}) \\ 
    & \mathbf{p}''(s)=2(\mathbf{c}_{2}-2\mathbf{c}_{1}+\mathbf{c}_{0}). \\
\end{split}
\label{eq:quadratic_bezier_derivation}
\end{equation}
\textcolor{black}{Each cross-section is approximated as a circle with the radius $r(s)$, and the corresponding tubular surface is defined as: }
\begin{equation}
\begin{split}
    \mathbf{S}(s, \phi) = \mathbf{p}(s) + r(s) \left[-\mathbf{N}(s)\cos{\phi} + \mathbf{B}(s)\sin{\phi} \right]
\end{split}
\end{equation}
\textcolor{black}{with $\phi \in \left[0, 2\pi \right]$.
Since a point on the tubular surface can be specified by $s$ and $\phi$, we compute the mesh vertices by discretizing the tubular surface. The mesh vertices are then defined by a set of points on the tubular surface with two additional points at both ends of the curve,}
\begin{equation}
    \mathcal{V} = \left\lbrace \mathbf{S}(s_i, \phi_i),  \mathbf{p}(0), \mathbf{p}(1) \; |\; i = 1,...,N_d\right\rbrace 
\end{equation}
\textcolor{black}{where $s_i, \phi_i$ are discrete points for surface vertices.
The example of reconstructed surface mesh is shown in Fig.~\ref{fig:bezier_c}.
During the optimization process, we adjust the mesh vertices by optimizing the radius of the cross sections $\Theta_{verts} := r(s)$, and the robot mesh is formed with adjusted vertices.}
\vspace{2mm}
\subsubsection{Mesh Reconstruction for Robot Manipulator}
\label{methods:rigid_robot}

A robot manipulator can generally be described as a chain of rigid links, and the 3D robot mesh is separated into a set of individual meshes of primitive shape. The number of individual meshes equals the number of rigid links and the meshes are connected by the rigid body transformations which indicate the position and rotation with respect to the robot base frame $\{b\}$. The transformation matrices $\mathbf{T}^b_n(q_1,...,q_n) \in SE(3)$ are parameterized by joint angles and can be computed from forward kinematics~\cite{dhparameters}. 

We initialize the set of individual meshes as primitive shapes (e.g. boxes or cylinders) with predefined edges, faces, and vertices $\mathcal{V}_{\mathtt{primitive}}$. Note that anything can be a primitive shape as long as there are only a few tunable parameters defining it, so a link shape template could be used. Regardless, during the optimization process, we adjust the robot mesh by estimating the offsets of each vertex:
\begin{equation}
\label{eq:vertex_offset}
    \mathbf{v} = \mathbf{v}_{\mathtt{primitive}} + \mathbf{v}_{\mathtt{offset}}
\end{equation}
where $\mathbf{v}_{\mathtt{primitive}} \in \mathcal{V}_{\mathtt{primitive}}$ is the vertex initialized with the primitive shape mesh and $\mathbf{v}_{\mathtt{offset}}$ is the corresponding offset vector. The set of offset vectors has the same number of elements as $\mathcal{V}_{\mathtt{primitive}}$, and is optimized at each iteration through gradient descent ($\Theta_{verts} := \mathcal{V}_{\mathtt{offset}}$). 

To compose the robot mesh $\mathcal{M}$ for pose estimation, each individual mesh needs to be connected by the forward kinematics and transformed to the camera frame. 
Let $\mathbf{v}^n \in \mathbb{R}^3$ be a vertex of the $n$-th link mesh, we transform the vertex to the camera frame as:
\begin{equation}
\label{eq:vertex_transform}
    \overline{\mathbf{v}}^c = \mathbf{T}^c_b \mathbf{T}^b_n(q_1,...,q_n) \overline{\mathbf{v}}^n
\end{equation}
where $\overline{\cdot}$ represents the homogeneous representation of a point (e.g. $\overline{\mathbf{v}} = [\mathbf{v}, 1]^T$). $\mathbf{T}^b_n(q_1,...,q_n)$ obtained from forward kinematics transforms mesh vertices to the robot base frame and $\mathbf{T}^c_b$ is the \textcolor{black}{robot-to-camera} transformation which is estimated using the Algorithm~\ref{alg:main_outline} ($\Theta_{state}:=\mathbf{T}^c_b$).

\subsection{Differentiable Rendering and Loss Functions}
\label{method:diff_rendering}
To render the image for robot mesh $\mathcal{M}$, we use the PyTorch3D differentiable render~\cite{pytorch3d} for silhouette rendering. We set up the silhouette renderer with a perspective camera and a \textit{SoftSilhouetteShader} which does not apply any lighting and shading. 
The differentiable renderer applies the rasterization algorithm which finds the mesh triangles that intersect each pixel and weights the influence according to the distance along the $z$-axis.
Finally, the \textit{SoftSilhouetteShader} computes pixel values of the rendered silhouette image using the sigmoid blending method~\cite{liu2019soft}.

\vspace{2mm}
\subsubsection{Objective Loss Functions for Shape Reconstruction}
\label{method:loss_shape}

To minimize the difference between the reconstructed silhouette image and the observed binary mask, the commonly used mask loss is applied. The mask loss computes the sum of the mean square error for every pixel,
\begin{equation}
        \mathcal{L}_{mask} = \sum_{i=0}^{H-1} \sum_{j=0}^{W-1} \left(\mathbb{S}(i,j) - \mathbb{M}^{ref}(i,j)\right)^2.
\label{eq:mask_loss}
\end{equation}
\noindent $H$ and $W$ is the image height and width, $\mathbb{S}$ is the rendered silhouette image and $\mathbb{M}^{ref}$ is the reference binary mask.

\begin{figure}[t!]
\subfloat[]{\includegraphics[width=0.31\linewidth]{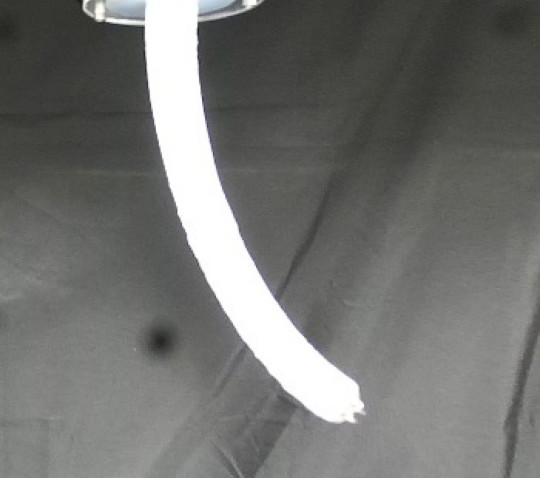}\label{fig:preprocess_a}}
\hfill
\subfloat[]{\includegraphics[width=0.31\linewidth]{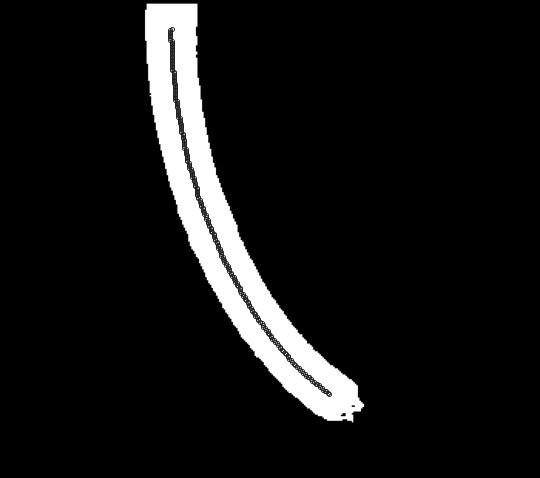}\label{fig:preprocess_b}}
\hfill
\subfloat[]{\includegraphics[width=0.31\linewidth]{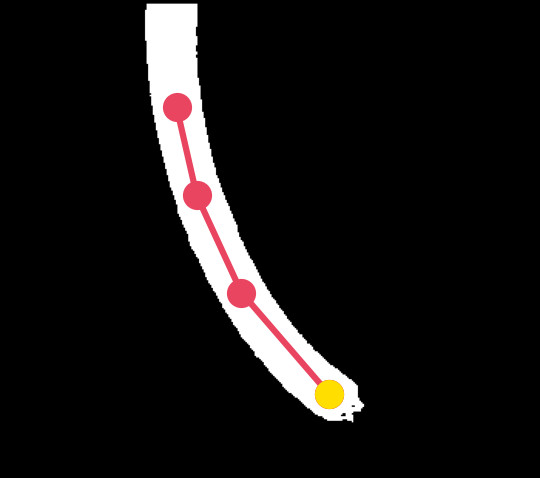}\label{fig:preprocess_c}}
\caption{The pre-processing of soft octopus arm images, with \protect\subref{fig:bezier_a}~the observed RGB image, \protect\subref{fig:bezier_b}~the reference binary mask with center-line and \protect\subref{fig:bezier_c}~the predefined keypoints and the endpoint (yellow).}
\label{fig:preprocessing_soft_arm}
\vspace{-0.14in}
\end{figure}

The mask loss will have non-informative gradients when there is no overlap (e.g. $\mathbb{S}(i,j) = 0$ but $\mathbb{M}^{ref}(i,j) = 1$). Therefore we use an additional keypoint loss to guide the optimization from local minima when the silhouettes do not overlap. The keypoints loss is defined as:
\begin{equation}
    \begin{split}
        & \mathcal{L}_{keypoint} = \sum_{i=1}^K \left\| \pi( \mathbf{p}_i )  - \mathbf{\hat{x}}_i \right\|_2 \\
    \end{split}
\end{equation}
where $K$ is the number of keypoints, $\mathbf{\hat{x}}_i$ is the $i$-th 2D keypoint extracted from center line of the reference mask $\mathbb{M}^{ref}$ as shown in Fig.~\ref{fig:preprocessing_soft_arm}, and $\mathbf{p}_i$ is the corresponding 3D keypoint on the B\'{e}zier curve. $\pi(\cdot )$ is the camera projection operator.
Finally, the shape reconstruction loss is defined as:
\begin{equation}
    \mathcal{L}_{shape} = \lambda_{mask} \mathcal{L}_{mask} + \lambda_{keypoint} \mathcal{L}_{keypoint} 
\end{equation}
with $\lambda_{mask},\lambda_{keypoint}$ as loss weights.

\vspace{2mm}
\subsubsection{Objective Loss Functions for Pose Estimation}
\label{method:loss_rigid}

\begin{figure}[t!]
\vspace{-2mm}
\subfloat[]{\includegraphics[height=0.37\linewidth]{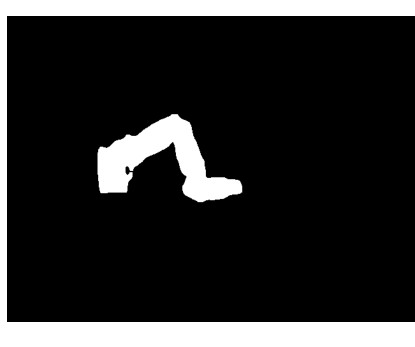}\label{fig:dist_map_a}}
\hfill
\subfloat[]{\includegraphics[height=0.37\linewidth]{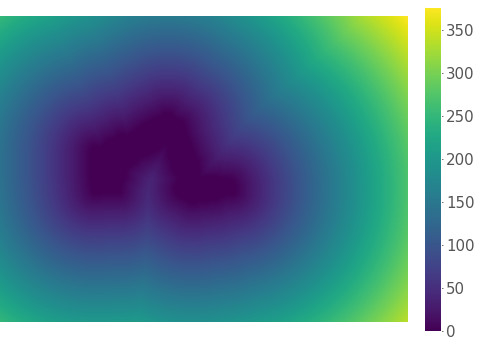}\label{fig:dist_map_b}}
\caption{Visualization of the distance map, with \protect\subref{fig:dist_map_a}~the reference binary mask and \protect\subref{fig:dist_map_b}~the corresponding distance map.}
\label{fig:dist_map}
\vspace{-0.14in}
\end{figure}

For robot pose estimation, poor initialization would hamper the performance of the optimization algorithm by converging to local minima.
In addition to the commonly used mask loss (Eq.~\ref{eq:mask_loss}), we propose distance loss and appearance loss to aid the optimization.
The distance loss utilizes the distance map to propagate the gradient information to the entire image. The distance map $\mathbb{D}^{ref}$ is defined as:
\begin{equation}
    \mathbb{D}^{ref}(i,j) = 
\begin{cases}
    0,              & \text{if } \mathbb{M}^{ref}(i,j)=1\\
    \frac{dist(i,j)}{\gamma},              & \text{otherwise}
\end{cases}
\end{equation}
where $dist(i,j)$ is the distance from the pixel $(i,j)$ to the closest pixel that has positive value 1, and $\gamma$ is a discount factor (Fig.~\ref{fig:dist_map}). We use the \textit{scikit-fmm} package\footnote{\url{https://pythonhosted.org/scikit-fmm}} which implements the fast marching method~\cite{sethian1999fast} for computing the distance map. The distance loss is then calculated as:
\begin{equation}
    \mathcal{L}_{dist} = \sum_{i=0}^{H-1} \sum_{j=0}^{W-1} \mathbb{S}(i,j) * \mathbb{D}^{ref}(i,j)
\end{equation}
Since the reconstructed mask should have the same appearance as the observed mask, we introduce the appearance loss to force them to have the same number of positive pixels:
\begin{equation}
        \mathcal{L}_{app} = \left\| \sum_{i=0}^{H-1} \sum_{j=0}^{W-1} \mathbb{S}(i,j) - \sum_{i=0}^{H-1} \sum_{j=0}^{W-1} \mathbb{M}^{ref}(i,j) \right\|
\end{equation}
The appearance loss is effective for preventing the robot pose from being too far or too close to the camera, through regulating the size of the rendered mask.
Finally, the objective loss for pose estimation is defined as:
\begin{equation}
    \mathcal{L}_{pose} = \lambda_{mask} \mathcal{L}_{mask} + \lambda_{dist} \mathcal{L}_{dist} + \lambda_{app} \mathcal{L}_{app}
\end{equation}
where $\lambda_{mask},\lambda_{dist},\lambda_{app}$ are weights for the loss functions.

\section{Experiments and Results}
\label{section:experiment}

\subsection{Datasets and Evaluation Metrics}


\textbf{Tendon-driven Octopus Arm dataset}. Our experimental evaluations are conducted on a physical prototype of tendon-driven octopus arm, visible in Fig.~\ref{fig:cover}. It consists of a tapered cylinder of Ecoflex 00-30 (Smooth-On, Inc., Macungie, PA, USA) of length $200$~mm, base and tip radius of $10$ and $6$~mm, respectively. It contains four channels for tendon actuation. The tendons are rigidly attached at the tip, and connected to spools actuated by harmonic drive motors at the base. The motors displace the tendons, leading to motions of the octopus arm. We collected the images using the ZED camera and the ground-truth robot shape is obtained with stereo reconstruction. For evaluation, we compare the center line of the reconstructed robot shape with the ground truth shape. The 2D and 3D errors of the center line are computed using the Euclidean distance of discrete points.



\begin{figure*}[t]
    \centering
    \subfloat[]{\includegraphics[height=0.4\linewidth]{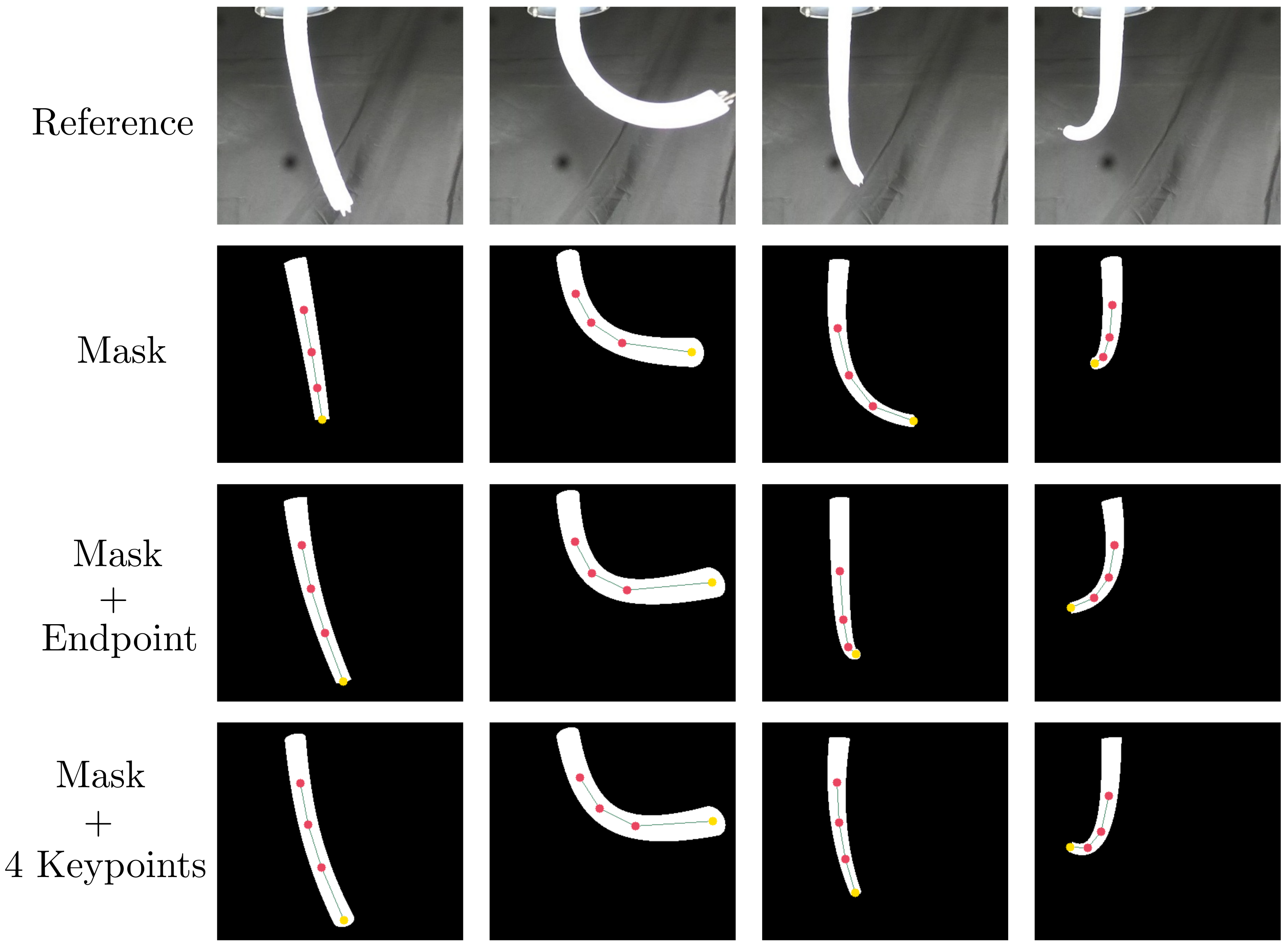}\label{fig6a}}
    \hfill \hfill
    \subfloat[]{\includegraphics[height=0.4\linewidth]{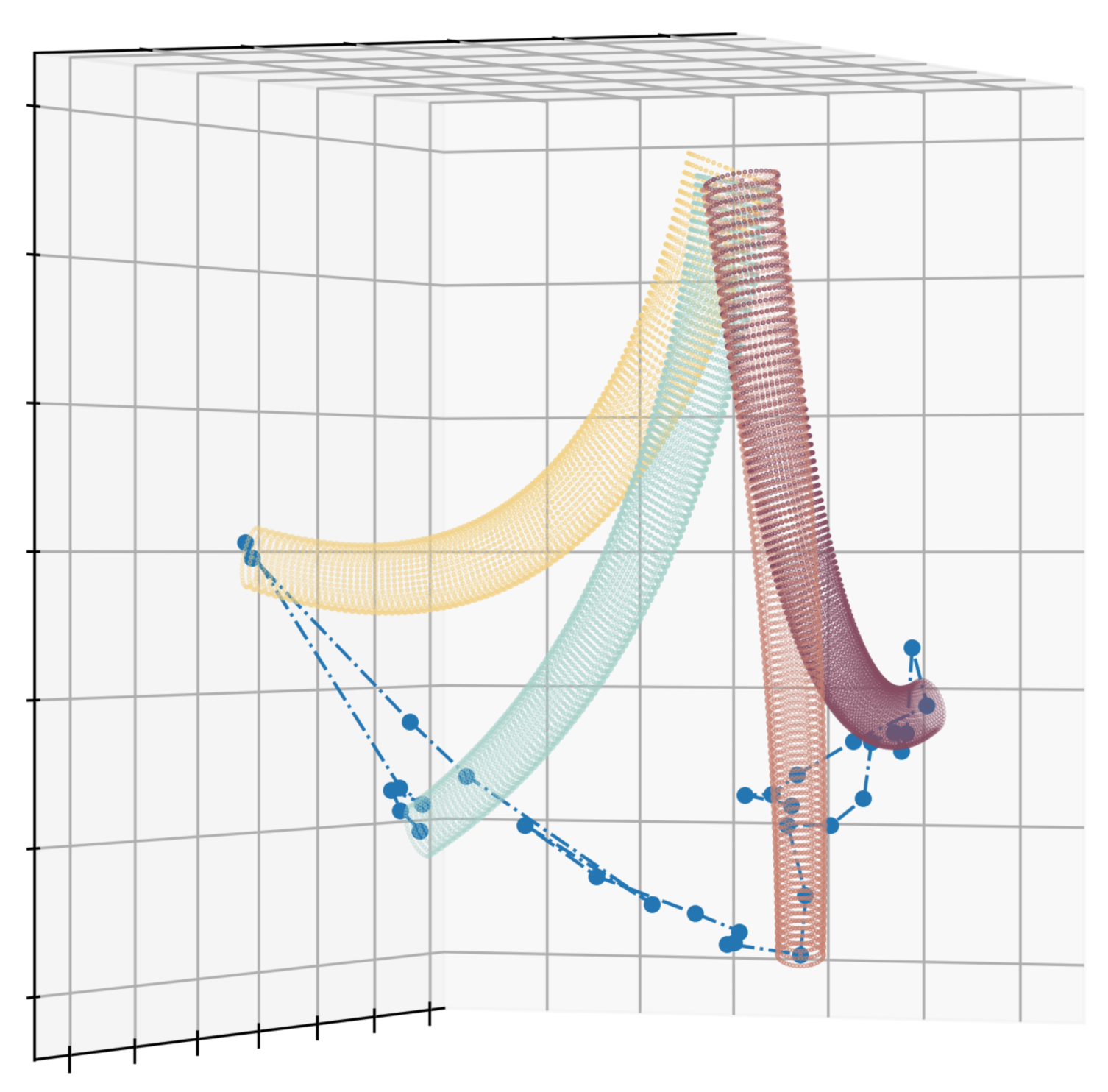}\label{fig6b}}
    \caption{Reconstruction results for the Octopus Arm dataset, with \protect\subref{fig6a}~the reference RGB images and shape reconstruction results for different losses are shown for 4 example frames that cover a large range of motion and \protect\subref{fig6b}~the reconstructed 3D robot shape of the picked frames and the entire robot trajectory.
    }
    \label{fig:soft_octopus_robot_final_result}
    \vspace{-0.2in}
\end{figure*}

\begin{table}[b!]
\setlength\tabcolsep{1em}
\centering
\caption{\label{tab:octupus_sim_errors} 2D and 3D Error (mean $e$ and standard deviation $\sigma$) of Shape Reconstruction on real octopus Arm Dataset.}
\begin{tabular}{lcccc}
\toprule
Losses & $e_{2D}$ (pixel) & $\sigma_{2D}$ & $e_{3D}$ & $\sigma_{3D}$\\
\midrule
Mask & 12.462 & 8.690 & 8.720 & 2.534 \\
Mask + endpoint & 3.898 & 2.216 & 7.299 & 3.900 \\
Mask + 4 keypoints & \bf{3.276} & \bf{0.785} & \bf{6.915} & \bf{2.096} \\
\bottomrule
\end{tabular}
\end{table}

\textbf{Baxter dataset}. The Baxter dataset from~\cite{lu2022pose} provides 100 image frames of 20 different robot poses. The ground-truth end-effector position in 2D and 3D are provided. This dataset includes the challenging scenarios where the robot manipulator is self-occluded. The Percentage of Correct Keypoints (PCK) metric of the end-effector will be calculated according to~\cite{lu2022pose}, where the end-effector position in the camera frame is calculated based on the estimated robot pose.



\subsection{Shape Reconstruction for Soft Continuum Robot}
\label{section:shape_reconstruction}

\subsubsection{Implementation details}
The RGB images are pre-processed to binary masks and the 2D center-line are extracted from the reference binary mask using the \textit{scikit-image} (\url{https://scikit-image.org}) package, which implements the fast skeletonization method~\cite{zhang1984fast}.
We \textcolor{black}{arbitrary} predefined 4 keypoints along the center-line for loss computation, as shown in Fig.~\ref{fig:preprocessing_soft_arm}.
\textcolor{black}{For computing the mesh vertices, $s$ is discretized to 100 and $\theta$ is discretized to 40 number of evenly spaced points. For the loss function, we set $\lambda_{mask}=1$ and $\lambda_{keypoint}=100$. We initialize the control points randomly but make sure the initialized mesh is within the camera frustum. The optimization loop is run for 200 iterations with a learning rate of 0.2.}

\vspace{2mm}
\subsubsection{Evaluation on the Octopus Arm Dataset}
We evaluate our shape reconstruction method with different loss functions described in Section~\ref{method:loss_shape}. 
For the keypoint loss, we experimented with only using the endpoint and using all 4 keypoints. 
We report the averaged 2D and 3D center-line error and the results are shown in Table~\ref{tab:octupus_sim_errors}.
\textcolor{black}{We can see that the error is dropped significantly by combining the mask loss and keypoint loss with only the endpoint. Considering more keypoints further improves our performance of shape reconstruction \textcolor{black}{as they provide more guidance for optimization}. The qualitative results are shown in the Fig.~\ref{fig:soft_octopus_robot_final_result}, where we show the rendered silhouette images of the reconstructed robot mesh (left) and, the reconstructed 3D robot shape, and the robot trajectory (right).}

\begin{table*}[t]
\centering
\caption{\label{exp:rigid_pck}Comparison of our methods with the state-of-the-art methods on robot pose estimation.}
\setlength\tabcolsep{0.9em}
\begin{tabular}{lcccccccc}
\toprule
\multirow{2}{*}{} & \multicolumn{4}{c}{PCK2D} & \multicolumn{4}{c}{PCK3D}\\ \cmidrule(lr){2-5} \cmidrule(lr){6-9}
                  & @50 pixel & @100 pixel & @150 pixel & @200 pixel & @100 mm & @200 mm & @300 mm & @400 mm \\ \midrule
DREAM~\cite{lee2020camera} & 0.33 & 0.52 & 0.62 & 0.64 & 0.32 & 0.43 & 0.54 & 0.66 \\
Optimized Keypoints~\cite{lu2022pose}& 0.69 & 0.88 & 0.93 & 0.95 & 0.47 & 0.74 & 0.86 & 0.90\\
Ours (box)          & 0.65 & \bf{0.94} & \bf{0.95} & 0.95 & \bf{0.8} & \bf{0.95} & 0.95 & 0.95\\
Ours (cylinder)     & \bf{0.80} & 0.91 & 0.93 & 0.95 & 0.71 & 0.93 & 0.94 & 0.95 \\
Ours (CAD)          & 0.74 & 0.90 & 0.94 & \bf{1.0} & 0.78 & 0.93 & \bf{0.97} & \bf{1.0} \\
\bottomrule
\end{tabular}
\end{table*}



\subsection{Pose Estimation for Robot Manipulator}
\label{section:pose_estimation}

\subsubsection{Implementation details}
To segment the robot from the background, we trained the DeepLabV3~\cite{chen2017rethinking} with 10K synthetic image data generated using the CoppeliaSim~\cite{rohmer2013v}. We applied Domain Randomization~\cite{tobin2017domain} so that the trained network can generalize to the real images.
We initialize the primitive shape meshes for each link with the length, width, and height that are described in the robot description file. The deformed vertices $\mathbf{v}_{\mathtt{deformed}}$ are initialized to zeros and the robot poses are initialized randomly but within the camera frustum.
For loss function, we set the weights as $\lambda_{mask}=1, \lambda_{dist}=1, \lambda_{app}=1$ and $\gamma=100$ when computing the distance map.
We optimize parameters for 500 iterations with a learning rate of 1e-2 for camera pose parameters and 1e-4 for the deformed vertices.

\vspace{2mm}
\subsubsection{Evaluation on Baxter Dataset}
We evaluate our method of robot pose estimation on the Baxter dataset and compare it against the state-of-the-art methods~\cite{lee2020camera, lu2022pose}, as shown in Table~\ref{exp:rigid_pck}. We applied our method with two different shape primitive meshes, the box and cylinder. We also report the performance of using the robot CAD model with differentiable rendering. 
The Percentage of Correct Keypoints (PCK, higher is better) results are reported for both 2D and 3D at different thresholds. 
The experimental results show that our method of using primitive shapes outperforms the state-of-the-art methods and achieves comparable performance with using the high-resolution robot CAD model.

\vspace{2mm}
\subsubsection{Ablation Study on Loss Functions}


\begin{figure}[t!]
\centering
\input{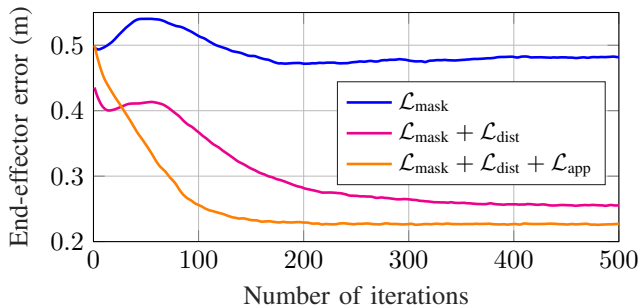}
\vspace{-6mm}
\caption{\label{fig:pose_loss_abalation} Ablation study of the loss functions for robot pose estimation. \textcolor{black}{Notice that the averaged error is large because of outliers, and 95\% of the estimates have less than 4 cm error as shown in Table~\ref{exp:rigid_pck}.}
}
\vspace{-0.28in}
\end{figure}

Here, we study the effectiveness of the loss functions proposed in Section~\ref{method:loss_rigid} using the Baxter dataset. We experiment with our robot pose estimation method (cylinder) with different loss combinations and calculate 3D end-effector error using the Euclidean Distance. We plot the average 3D end-effector error at each iteration in Fig.~\ref{fig:pose_loss_abalation}. With only the mask loss, the algorithm suffers from bad initialization and cannot converge robustly. The distance loss helps the convergence by propagating the gradient information to every image pixel.
Finally, by combining all three losses, we achieve a more robust convergence for robot pose estimation.

\section{Conclusion}
In this paper, we demonstrate the capability of measuring robot pose and configuration state parameters directly from a camera, as shown in Fig.~\ref{fig:cover}. The method works via the technique of differentiable rendering, and can be effective both in rigid-link robot manipulators as well as soft continuum robots.
we show that several definitions for optimization losses are useful to overcome the local minima when applying differentiable rendering to the objective of robot pose estimation.
We evaluated our method on relatively unstructured environments of continuum and rigid robots showing its efficacy in pose estimation. Ultimately, this work helps to enable robot state estimation and tracking \textit{in the wild}, with greater opportunities in useful dataset curation, behavioral cloning and visual learning.

\balance
\bibliographystyle{ieeetr}
\bibliography{references}

\end{document}